\documentclass[conference]{IEEEtran}
\IEEEoverridecommandlockouts
\usepackage{cite}
\usepackage{amsmath,amssymb,amsfonts}
\usepackage{algorithmic}
\usepackage{stfloats}
\usepackage{graphicx}
\usepackage{subcaption}
\usepackage{graphicx}
\usepackage{booktabs}
\usepackage{textcomp}
\usepackage{xcolor}
\def\BibTeX{{\rm B\kern-.05em{\sc i\kern-.025em b}\kern-.08em
    T\kern-.1667em\lower.7ex\hbox{E}\kern-.125emX}}
\begin{document}

\title{Vision-Guided Morphing Quadcopter for Multi-Geometry Payload Transport through Narrow Passages}

\author{
\IEEEauthorblockN{Aashish Sahu, Shriram Hari, and R. Prasanth Kumar}
\IEEEauthorblockA{
\textit{Department of Mechanical and Aerospace Engineering}\\
\textit{Indian Institute of Technology Hyderabad}\\
Hyderabad, India\\
\{me22resch11011, me22btech11023\}@iith.ac.in, rpkumar@mae.iith.ac.in}
}
\maketitle

\begin{abstract} Aerial payload transport using multirotor unmanned aerial vehicles is challenging because payload geometry, contact interaction, grasp stability, flight control, and narrow-passage traversal are strongly coupled during pickup and transport. Conventional quadcopters with fixed landing structures or object-specific grippers often cannot adapt their footprint or grasp geometry when the payload shape or passage width changes. This paper presents a vision-guided morphing quadcopter for multi-geometry payload transport through narrow passages. The proposed platform uses four hybrid arm--leg structures that function as both landing supports and grasping members. A centrally placed actuator drives a tendon-based morphing mechanism, enabling all four arms to synchronously retract or expand for object grasping, footprint reduction, and post-transport release. Onboard vision estimates the payload geometry and passage width, while endpoint force feedback is used to confirm grasp contact during payload engagement. A phase-wise mission planner, PID-based flight stabilization, and morphology-adaptive grasp controller are implemented in a MuJoCo simulation environment. The framework is evaluated using box, cylindrical, and spherical payloads, representing flat-faced, rolling-curved, and fully curved contact conditions. Across the three cases, the simulated system completes the pickup--transport--release sequence with a maximum RMS position error of 0.31~m, final drop-zone error below 0.18~m, compact grasp footprint of 0.09--0.21~m$^2$, and footprint reduction of 75.0--89.7\%. The results demonstrate that a single-actuator morphing quadcopter can adapt its grasp footprint for different payload geometries while reducing its effective width for narrow-passage transport. \end{abstract} 

\begin{IEEEkeywords} Morphing quadcopter, aerial grasping, payload transport, narrow-passage traversal, vision-guided grasping, tendon-driven mechanism, force feedback, MuJoCo simulation. \end{IEEEkeywords}

\section{Introduction}
Multirotor unmanned aerial vehicles (UAVs) have evolved from passive sensing platforms into active aerial robotic systems capable of inspection, payload transport, grasping, perching, and physical interaction. Their hovering capability, maneuverability, and access to cluttered or human-inaccessible environments make them suitable for indoor logistics, disaster response, industrial inspection, and constrained-space robotic missions. However, aerial payload grasping remains challenging because flight stability, payload contact, grasp force, center-of-mass variation, rotor disturbances, and navigation constraints are strongly coupled during pickup and transport \cite{ruggiero2018aerial,ladig2021aerial,meng2022aerial}. Unlike ground manipulators, aerial robots must maintain stable flight while simultaneously interacting with objects and compensating for payload-induced disturbances.

Existing aerial grasping systems commonly use rigid grippers, robotic arms, hooks, suction devices, compliant grippers, or other task-specific end-effectors. Vision-based aerial grasping improves object localization and alignment \cite{lin2019vision}, while compliant and soft grippers improve tolerance to contact uncertainty during payload engagement \cite{fishman2021soft,ubellacker2023aggressive}. Nevertheless, many approaches remain dependent on known grasp points, specific payload geometries, or dedicated gripper mechanisms. This limits their ability to handle different object shapes using the same platform. A box provides flat faces and edge contacts, a cylinder introduces rolling contact, and a sphere provides no stable planar grasping surface. Therefore, a fixed gripper may not reliably transport multiple payload geometries without redesign, additional actuation, or complex grasp replanning.

A second limitation of conventional quadcopters is their fixed physical footprint. In narrow indoor passages, such as windows, corridors, door frames, collapsed structures, and industrial openings, the full rotor-to-rotor span can restrict safe traversal. Morphing and reconfigurable quadrotors address this issue by changing their frame geometry during flight and have shown potential for narrow-gap traversal, obstacle-rich navigation, and compact flight \cite{xing2024morphing,shiferaw2021morphed,daadi2026reconfigurable}. However, most morphing-UAV studies focus mainly on navigation, whereas aerial grasping studies generally use a separate gripper for object pickup. The integrated problem of multi-geometry payload grasping and narrow-passage transport using the same morphology-adaptive quadcopter structure remains less explored.

Geometric caging offers a useful grasping principle because the object is constrained inside a controlled contact envelope instead of relying only on exact force-closure contact points \cite{makita2017caging,kim2019caging}. This is attractive for aerial transport, where contact uncertainty, rotor disturbance, and payload motion can make precise grasping difficult. However, aerial caging requires a lightweight and synchronized mechanism that does not increase actuation complexity. A single-actuator morphing structure that can serve as both landing support and grasping mechanism is therefore suitable for compact aerial manipulation.

To address these gaps, this paper presents a vision-guided morphing quadcopter for multi-geometry payload transport through narrow passages. The proposed quadcopter uses four hybrid arm--leg structures that function as landing supports during touchdown and as grasping members during payload engagement. A centrally located motor drives a tendon/thread-based mechanism, enabling all four arms to retract or expand synchronously using one actuator. During free flight, the vehicle remains in an open configuration for stability and rotor clearance. During grasping, the central motor retracts the tendon mechanism so that the hybrid arm--leg members move inward and form a compact caging-style grasp. During narrow-passage transport, the same morphology change reduces the effective vehicle footprint.

The proposed framework integrates onboard vision, endpoint force feedback, phase-wise trajectory generation, morphology-adaptive grasping, and PID-based flight stabilization in MuJoCo \cite{todorov2012mujoco}. The camera estimates payload size, shape, and passage width, while force sensors at the arm endpoints verify contact during grasping. The system is evaluated using box, cylindrical, and spherical payloads, representing flat-faced, rolling-curved, and fully curved contact conditions. By combining single-actuator morphology adaptation, hybrid arm--leg grasping, vision-guided alignment, and force-aware payload engagement, the proposed approach provides a compact solution for aerial payload transport through narrow passages.

\section{System Architecture}
\label{sec:system_architecture}

\begin{figure*}[t]
    \centering
    \includegraphics[width=\textwidth]{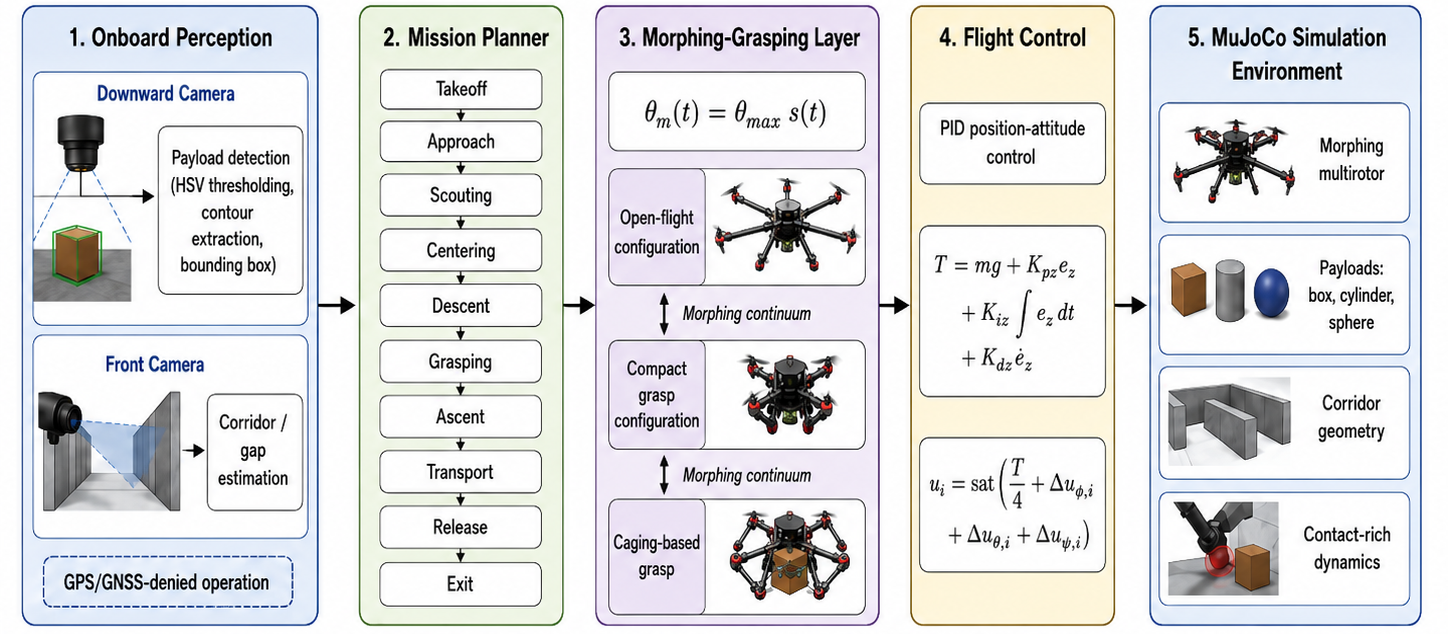}
    \caption{System architecture of the proposed vision-guided morphing quadcopter for multi-geometry payload grasping and narrow-passage transport. The framework integrates onboard perception, phase-wise mission planning, single-actuator tendon-driven morphology adaptation, endpoint force feedback, and closed-loop flight control in MuJoCo.}
    \label{fig:system_architecture}
\end{figure*}

The proposed system is a morphology-adaptive quadcopter equipped with four hybrid arm--leg structures. These members act as landing supports during touchdown and as grasping members during payload engagement. Unlike conventional aerial manipulators that require separate grippers or multiple actuators, the proposed design uses a centrally located motor to drive a tendon/thread-based mechanism. Rotation of the central motor synchronously retracts or expands all four arm--leg members, allowing the quadcopter to change its effective footprint during free flight, grasping, narrow-passage transport, and release.

The overall framework consists of five connected layers: MuJoCo simulation, onboard perception, mission planning, morphology-adaptive grasping, and flight control, as shown in Fig.~\ref{fig:system_architecture}. The MuJoCo environment models the quadcopter body, four rotors, central morphing actuator, hybrid arm--leg members, contact interaction, corridor geometry, and three payload geometries: box, cylinder, and sphere. For all payloads, the UAV follows the same task sequence: takeoff, approach, alignment, descent, grasping, grasp transport through the narrow passage, release, and final exit.

The perception layer estimates both payload and environment information. A downward-facing camera detects the payload using color-based segmentation, contour extraction, and bounding-box estimation, while the extracted contour provides approximate object size and shape information. A front-facing camera estimates the available passage width from the detected gap between obstacle boundaries. This enables the UAV to decide whether to remain in the open-flight configuration or reduce its footprint before entering the narrow passage.

The morphology-adaptive grasping layer converts perception and force information into a central motor command. During free flight, the arm--leg members remain open to preserve rotor clearance and stability. During payload engagement, the central actuator retracts the tendon/thread mechanism so that all four arm--leg members move inward and form a compact caging-style grasp around the payload. Endpoint force sensors near the arm tips confirm contact and verify sufficient payload confinement before transport. During release, the motor command is reversed and the arm--leg members return to the open configuration.

The morphing command is represented as
\begin{equation}
    \theta_m(t)=\theta_{\max}s(t),
\end{equation}
where $\theta_m(t)$ is the commanded morphing angle, $\theta_{\max}$ is the maximum morphing angle, and $s(t)\in[0,1]$ is the mission-phase-dependent transition variable. When $s(t)=0$, the quadcopter remains open; as $s(t)$ increases, the tendon mechanism retracts the arm--leg members and reduces the footprint for grasping and narrow-passage transport. During release, $s(t)$ decreases and the structure expands again.

The effective arm radius and grasp-footprint area are used to quantify morphology adaptation:
\begin{equation}
    r_{\mathrm{eff}}(t)=\frac{1}{N}\sum_{i=1}^{N}\|p_{g_i}(t)-p_{\mathrm{com}}(t)\|,
\end{equation}
\begin{equation}
    A_g(t)=\mathrm{Area}\left(\mathrm{ConvHull}\{p_{g_1}^{xy},p_{g_2}^{xy},...,p_{g_N}^{xy}\}\right).
\end{equation}
Here, $p_{g_i}$ is the position of the $i$th grasping point, $p_{\mathrm{com}}$ is the drone center of mass, and $N$ is the number of grasping points. A reduction in $r_{\mathrm{eff}}$ and $A_g$ indicates compact grasping morphology, which supports both payload confinement and footprint reduction during narrow-passage transport.

\begin{table}[t]
\centering
\caption{Key simulation and system parameters.}
\label{tab:sim_params}
\begin{tabular}{lc}
\hline
Parameter & Value \\
\hline
Simulation platform & MuJoCo \\
Vehicle type & Morphing quadcopter \\
Morphing mechanism & Central motor with tendon/thread drive \\
Hybrid members & Four arm--leg structures \\
Payload geometries & Box, cylinder, sphere \\
Pickup location, $x_o$ & 2.0 m \\
Drop-zone location, $x_g$ & 5.0 m \\
Initial height & 0.5 m \\
Nominal flight height & 1.5 m \\
Grasp height & 0.7--0.9 m \\
Maximum motor force, $u_{\max}$ & 15 N \\
Morphing command range, $\theta_{\max}$ & 1.7--2.0 rad \\
Mission duration & 45 s \\
Perception method & Camera-based contour detection \\
Contact feedback & Endpoint force sensing \\
Control method & PID position--attitude control \\
\hline
\end{tabular}
\end{table}

\section{Methodology and Control Design}
\label{sec:methodology}

\begin{figure*}[t]
    \centering
    \includegraphics[width=\textwidth]{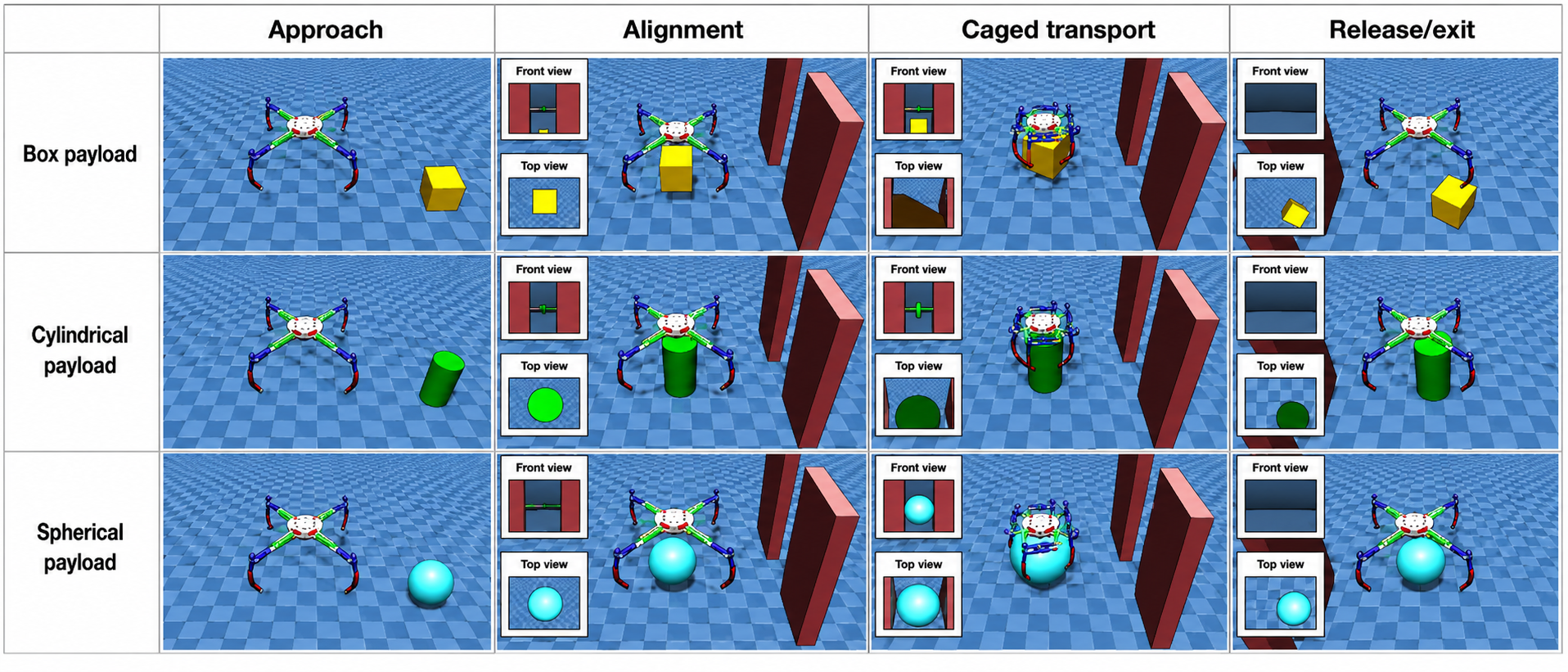}
    \caption{Multi-stage MuJoCo visualization of the proposed vision-guided morphing quadcopter during multi-geometry payload transport. The columns show approach, alignment, grasp transport, and release/exit phases for box, cylindrical, and spherical payloads.}
    \label{fig:multishape_grasp_transport}
\end{figure*}

The objective is to transport a payload from the pickup location to the drop zone while adapting the quadcopter morphology according to payload geometry and passage width. The same control framework is used for box, cylindrical, and spherical payloads. Let the UAV position be \(p_d=[x_d,y_d,z_d]^T\), and let the reference position generated by the mission planner be \(p_r=[x_r,y_r,z_r]^T\). The position-tracking error is defined as
\begin{equation}
    e_p(t)=p_r(t)-p_d(t).
\end{equation}

The mission planner generates phase-wise reference commands for takeoff, payload approach, visual alignment, descent, grasping, grasp transport, release, and final exit. During the scouting and alignment phase, the UAV performs a small circular search motion above the payload:
\begin{equation}
    x_r=x_o+r_s\sin(\omega_s t), \qquad
    y_r=y_o+r_s\cos(\omega_s t),
\end{equation}
where \(x_o\) and \(y_o\) denote the estimated payload location, and \(r_s\) and \(\omega_s\) are the scouting radius and angular rate, respectively. After localization, the UAV centers above the payload and descends to the required grasping height. This phase-wise trajectory reduces abrupt motion near the object and improves alignment before contact.

The onboard perception module estimates payload size, payload shape, and available passage width from camera feedback. The payload contour is extracted from the downward-facing camera image, and the approximate object width \(W_o\) is obtained from the bounding box. The front-facing camera estimates the passage width \(W_p\) from the visible gap between obstacle boundaries. Based on these measurements, the desired effective footprint is selected as
\begin{equation}
    W_d=\min\left(W_{\mathrm{open}},\, W_p-2\delta_p\right),
\end{equation}
where \(W_{\mathrm{open}}\) is the open-flight footprint and \(\delta_p\) is the safety margin from the passage boundary. The desired grasp envelope is selected as
\begin{equation}
    W_g = W_o + 2\delta_g,
\end{equation}
where \(\delta_g\) is the grasping clearance used to avoid excessive squeezing during arm closure.

The morphology command is implemented using a single central motor connected to the four hybrid arm--leg members through a tendon/thread-driven mechanism. The phase-dependent morphing command defined in the system architecture is used to retract or expand the arm--leg members. During free flight, the quadcopter remains in the open configuration. During grasping, the central motor retracts the tendon mechanism, causing all four arm--leg members to move inward synchronously and form a compact caging-style grasp around the payload. During narrow-passage transport, the compact morphology reduces the effective vehicle footprint. During release, the motor command is reversed to expand the arms.

Endpoint force sensors mounted at the arm tips are used as a contact-verification criterion during payload engagement. Let \(F_i(t)\) denote the measured contact force at the \(i\)th arm endpoint. The total grasp contact force is estimated as
\begin{equation}
    F_g(t)=\sum_{i=1}^{N_c}F_i(t),
\end{equation}
where \(N_c\) is the number of active contact points. A grasp is considered valid when the estimated contact force lies within a safe operating range:
\begin{equation}
    F_{\min} \leq F_g(t) \leq F_{\max}.
\end{equation}
Here, \(F_{\min}\) is the minimum force required to maintain payload confinement, and \(F_{\max}\) limits excessive compression of the payload. For a payload of mass \(m_p\), a conservative estimate of the minimum holding force is given by
\begin{equation}
    F_{\min} \geq \frac{m_p g}{\mu N_c},
\end{equation}
where \(\mu\) is the contact friction coefficient and \(g\) is the gravitational acceleration. The grasp-force margin is defined as
\begin{equation}
    M_F = F_g - F_{\min}.
\end{equation}
A positive value of \(M_F\) indicates that the contact condition is sufficient for payload confinement. In the present simulation, this force condition is used together with vision-based alignment as a logical grasp-confirmation criterion, rather than as a separate force-optimization objective.

Flight stabilization is achieved using a PID-based position--attitude controller. The altitude controller computes the total thrust demand as
\begin{equation}
    T=mg+K_{pz}e_z+K_{iz}\int e_z dt+K_{dz}\dot{e}_z,
\end{equation}
where \(m\) is the system mass and \(e_z=z_r-z_d\) is the altitude error. Lateral position errors are converted into desired roll and pitch commands, while attitude corrections are distributed among the four rotors. The command for the \(i\)th rotor is given by
\begin{equation}
    u_i=\mathrm{sat}\left(\frac{T}{4}+\Delta u_{\phi,i}+\Delta u_{\theta,i}+\Delta u_{\psi,i}\right),
\end{equation}
where \(\Delta u_{\phi,i}\), \(\Delta u_{\theta,i}\), and \(\Delta u_{\psi,i}\) are the roll, pitch, and yaw correction terms, respectively, and \(\mathrm{sat}(\cdot)\) limits the motor command within the allowable range.

The performance is evaluated using trajectory tracking, final drop-zone error, arm-radius reduction, grasp-footprint reduction, narrow-passage clearance, and \(X\)--\(Z\) flight path. The RMS tracking error is computed as
\begin{equation}
e_{\mathrm{rms}} =
\sqrt{\frac{1}{N}\sum_{k=1}^{N}
\left[(x_r(k)-x_d(k))^2+(z_r(k)-z_d(k))^2\right]} .
\end{equation}
The footprint reduction and arm-radius reduction are defined as
\begin{equation}
R_A =
\frac{A_{\mathrm{open}}-A_{\mathrm{compact}}}{A_{\mathrm{open}}}\times 100,
\end{equation}
\begin{equation}
R_r =
\frac{r_{\mathrm{open}}-r_{\mathrm{compact}}}{r_{\mathrm{open}}}\times 100.
\end{equation}
The final drop-zone error is computed as
\begin{equation}
    e_{\mathrm{drop}}=\sqrt{(x_p-x_g)^2+(z_p-z_g)^2},
\end{equation}
where \(x_p\) and \(z_p\) are the payload coordinates after release, and \(x_g\) and \(z_g\) are the desired drop-zone coordinates. A trial is considered successful if the UAV localizes the payload, forms a compact grasp, transports the object through the narrow passage, releases it within the goal region, and exits without loss of stability.

\section{Simulation Results and Discussion}
\label{sec:results}

The proposed vision-guided morphing quadcopter was evaluated in MuJoCo for multi-geometry payload transport through narrow passages. Three payload geometries were considered: box, cylinder, and sphere, representing flat-faced, rolling-curved, and fully curved contact conditions, respectively. In all cases, the same mission sequence was executed: takeoff, approach, visual alignment, descent, morphology-adaptive grasping, grasp transport through the constrained passage, release at the drop zone, and final exit. The payload was initially placed near \(x=2\)~m, and the drop zone was located near \(x=5\)~m.

Fig.~\ref{fig:multishape_grasp_transport} shows the stage-wise MuJoCo visualization of the task. During approach, the quadcopter remains in an open configuration to preserve rotor clearance and flight stability. During grasping, the central motor actuates the tendon/thread mechanism, causing the four hybrid arm--leg members to retract synchronously and form a compact caging-style grasp around the payload. This compact morphology reduces the effective vehicle footprint and enables grasp transport through the narrow passage. During release, the motor command is reversed, the arms expand, and the payload is released at the goal region.

\begin{figure*}[t]
    \centering
    \includegraphics[width=0.95\textwidth]{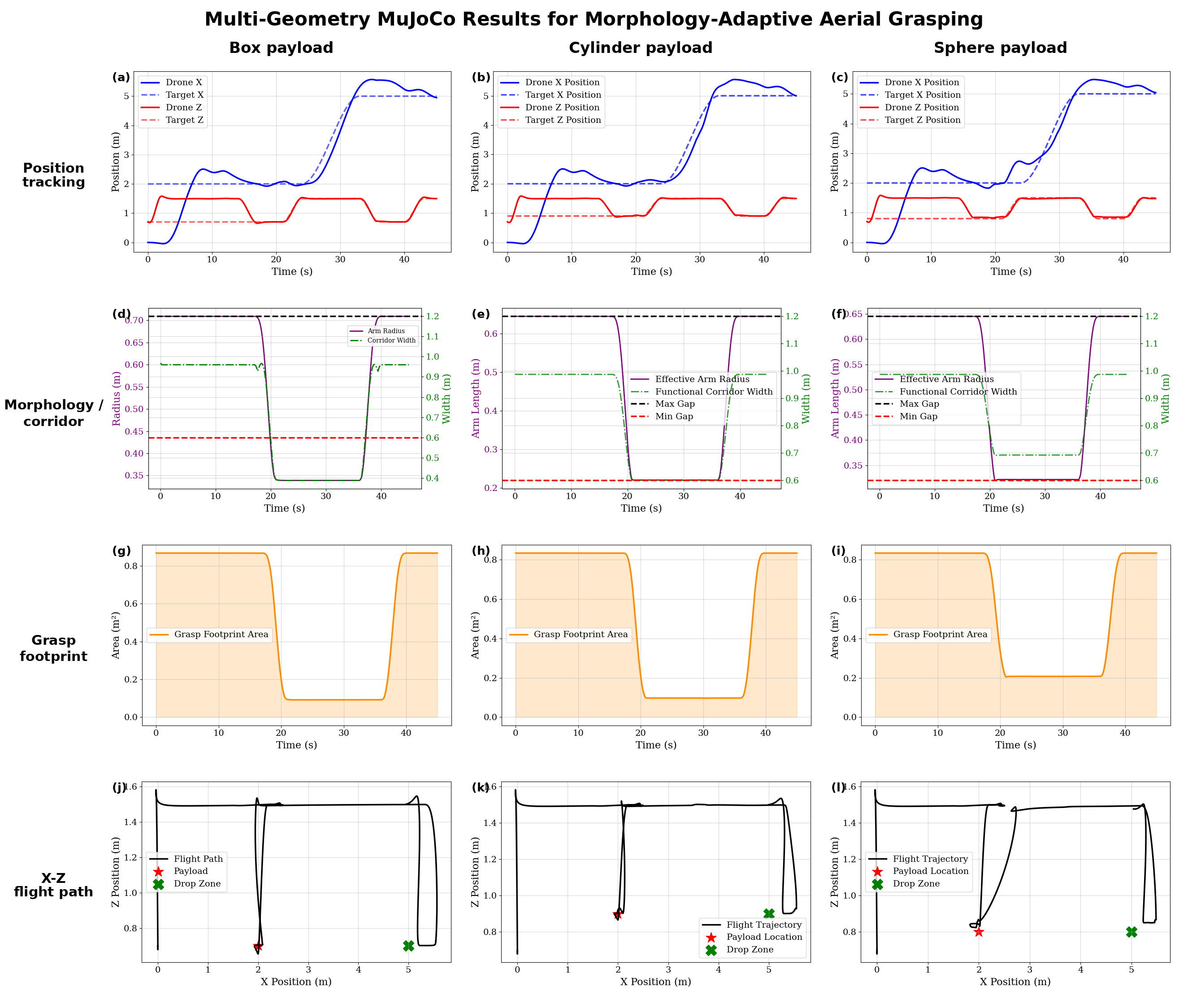}
    \caption{Overall MuJoCo simulation results for the proposed vision-guided morphing quadcopter. Rows show position tracking, morphology adaptation, grasp-footprint modulation, and \(X\)--\(Z\) flight path. Columns show box, cylindrical, and spherical payloads.}
    \label{fig:combined_results}
\end{figure*}

Fig.~\ref{fig:combined_results} presents the simulation responses for the three payloads, including position tracking, morphology adaptation, grasp-footprint modulation, and \(X\)--\(Z\) flight path. The UAV follows the commanded \(X\)- and \(Z\)-position references during approach, descent, grasping, transport, release, and final ascent. A small transient overshoot appears during grasp transport because payload contact, morphology change, and inertial coupling occur simultaneously. However, the vehicle remains stable and completes the pickup--transport--release sequence for all payload geometries.

Table~\ref{tab:quant_results} summarizes the quantitative performance metrics. The proposed framework completes the task for all payloads with a maximum RMS position error of 0.31~m and final drop-zone error below 0.18~m. The grasp-footprint area decreases from 0.84--0.87~m\(^2\) in the open configuration to 0.09--0.21~m\(^2\) in the compact grasping configuration, corresponding to a footprint reduction of 75.0--89.7\%. This confirms that the single central actuator actively changes the grasp geometry and effective vehicle footprint, instead of operating as a fixed-frame quadcopter.

\begin{table}[t]
\centering
\caption{Quantitative performance summary for vision-guided morphology-adaptive payload transport.}
\label{tab:quant_results}
\begin{tabular}{lccc}
\hline
Metric & Box & Cylinder & Sphere \\
\hline
Task completion & Yes & Yes & Yes \\
Max $X$ overshoot (m) & 0.54 & 0.50 & 0.56 \\
Max $Z$ overshoot (m) & 0.08 & 0.07 & 0.09 \\
RMS position error (m) & 0.29 & 0.27 & 0.31 \\
Final drop error (m) & 0.14 & 0.16 & 0.18 \\
Open arm radius (m) & 0.65 & 0.70 & 0.64 \\
Compact arm radius (m) & 0.22 & 0.34 & 0.32 \\
Arm-radius reduction (\%) & 66.2 & 51.4 & 50.0 \\
Open footprint area (m$^2$) & 0.84 & 0.87 & 0.84 \\
Minimum footprint area (m$^2$) & 0.09 & 0.09 & 0.21 \\
Footprint reduction (\%) & 89.3 & 89.7 & 75.0 \\
Minimum corridor width (m) & 0.60 & 0.60 & 0.69 \\
\hline
\end{tabular}
\end{table}

The morphology and footprint responses confirm the transition from open-flight morphology to compact grasping morphology during payload engagement and narrow-passage transport. After release, the grasp footprint increases again as the UAV returns to its open configuration. The \(X\)--\(Z\) flight paths verify the complete task behavior, including descent near the payload, ascent after grasping, forward transport, descent for release, and final exit.

The endpoint force-feedback condition was used as a grasp-confirmation criterion before initiating the transport phase. In this study, the force condition verifies contact establishment and prevents premature transport before payload engagement. The box payload provides stable face and edge contact, the cylindrical payload requires radial closure to reduce rolling tendency, and the spherical payload requires a wider compact grasp envelope because it has no flat contact surface. Despite these different contact conditions, the same hybrid arm--leg structure and single-actuator morphing strategy transports all three payloads. Overall, the results demonstrate that landing support, payload grasping, and footprint reduction can be integrated into one morphology-adaptive mechanism for constrained payload transport.

\section{Conclusion and Future Work}
\label{sec:conclusion}

This paper presented a vision-guided morphing quadcopter for multi-geometry payload transport through narrow passages. The proposed platform uses four hybrid arm--leg structures that function as both landing supports and grasping members. A single central actuator drives a tendon/thread-based mechanism to synchronously retract or expand all four arms, enabling payload grasping, footprint reduction, narrow-passage transport, and payload release using the same mechanism.

The framework was evaluated in MuJoCo using box, cylindrical, and spherical payloads. The simulation results showed successful pickup--transport--release performance for all three payload geometries. The maximum RMS position error was 0.31~m, the final drop-zone error remained below 0.18~m, and the grasp-footprint reduction reached 75.0--89.7\%. These results indicate that the proposed single-actuator morphology-adaptive design can provide shape-tolerant payload confinement while reducing the effective vehicle footprint for constrained-space transport.

The proposed approach offers a compact alternative to object-specific aerial grippers and fixed-frame quadcopters for indoor logistics, narrow-passage payload transfer, and autonomous object deployment. Future work will focus on depth-based perception, contact-aware control, improved force regulation, disturbance modeling, and experimental validation using a physical morphing-quadcopter prototype with real payloads of different shapes, masses, and surface properties.

\bibliographystyle{IEEEtran}
\bibliography{ref}

\end{document}